# Self-Attention with State-Object Weighted Combination for Compositional Zero Shot Learning

Cheng-Hong Chang and Pei-Hsuan Tsai,

*Abstract*—Currently, the application of object recognition has become prevalent across various industries, agriculture, and other domains. However, most existing applications are limited to identifying objects alone, without considering their associated states. The ability to recognize both the state and object simultaneously remains less common. One approach to address this is by treating "state and object" as a single category during training. However, this approach poses challenges in data collection and training since it requires comprehensive data for all possible combinations. Compositional Zero-shot Learning (CZSL) emerges as a viable solution by treating the state and object as distinct categories during training. CZSL facilitates the identification of novel compositions even in the absence of data for every conceivable combination. The current state-of-the-art method, KG-SP, addresses this issue by training distinct classifiers for states and objects, while leveraging a semantic model to evaluate the plausibility of composed compositions. However, KG-SP's accuracy in state and object recognition can be further improved, and it fails to consider the weighting of states and objects during composition. In this study, we propose SASOW, an enhancement of KG-SP that considers the weighting of states and objects while improving composition recognition accuracy. First, we introduce self-attention mechanisms into the classifiers for states and objects, leading to enhanced accuracy in recognizing both. Additionally, we incorporate the weighting of states and objects during composition to generate more reasonable and accurate compositions. Our validation process involves testing SASOW on three established benchmark datasets. Experimental outcomes affirm the effectiveness of SASOW, as it exhibits competitive performance. When compared against the state-of-the-art OW-CZSL approach, KG-SP, SASOW showcases improvements of 2.1%, 1.7%, and 0.4% in terms of accuracy for unseen compositions across the MIT-States, UT Zappos, and C-GQA datasets, respectively.

## I. INTRODUCTION

State refers to a semantic depiction that succinctly conveys the attributes inherent to an object [22, 23], encompassing characteristics like its material, color, style, shape, function, and more. These states provide a means to comprehend the particular aspects in which an object has undergone alteration, be it in terms of physical or chemical attributes. At present, the majority of object recognition technologies are limited to identifying the object itself, without accounting for any transformations it may have undergone. The absence of these contextual states can result in an inability to discern the intended purpose of the object. To illustrate, consider the scenario of having two photographs depicting tomatoes—one portraying a fresh tomato, and the other showing a moldy one. In this case, conventional object recognition systems would merely classify both images as depicting tomatoes, without the capability to distinguish between the two varying conditions.

The emergence of state-object recognition avoids this situation. State-object recognition is applicable in several fields. In manufacturing, it is possible to determine which flaws constitute inferior components. It is also feasible to determine whether the manufacturing procedure is correct by observing the product assembly process. In agriculture, it is possible to identify which crops are ripe enough to harvest. These applications will reduce a lot of manual judgment and improve the production efficiency.

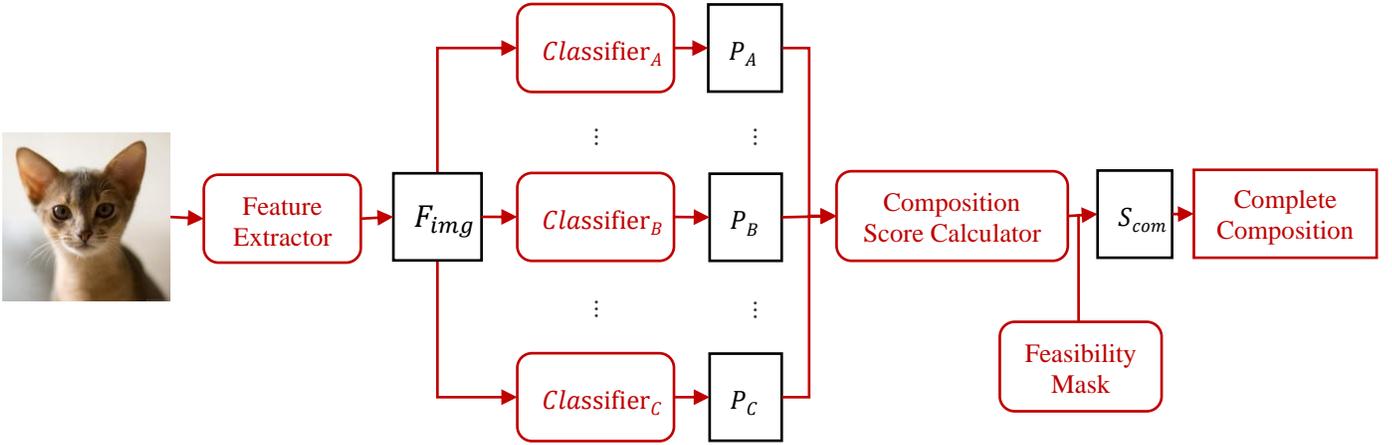

**Fig. 1. The recognition process of OW-CZSL is depicted in the diagram, where the red-colored components highlight the improvements made in this study. Specifically, in the classifier section, the concept of self-attention is incorporated. Additionally, the Composition score calculator incorporates the notion of weighted combination.**

While treating state-object as a single annotation can enable the training of models for such applications, it necessitates a large amount of compositional training data and fails to recognize compositions not included in the training data. Expanding the training dataset is necessary to identify new compositions. State-object recognition [23], on the other hand, addresses this issue by considering compositions as annotations of both states and objects. By leveraging the learned characteristics of states and objects during training, it becomes possible to discern combinations not present in the training data. According to the number of models employed, state-object recognition can be classified into the following three methods: The first method, referred to as *Simple Compositional Learning* (SCL), employs two identical model networks to identify states and objects, separately. SCL is simple to implement and compatible with common training data set. However, SCL may results in no existing match of state and object, such as melted dog because SCL ignores the interaction between states and objects. The second method is *Compositional Zero Shot Learning* (CZSL), which aims to recognize objects and their states using a single model. CZSL addresses the limitations of SCL by modeling state-object transformations and extending these transformations to compositions not seen during training. Various modeling methods can be employed in CZSL, such as compositional classifiers [1] or shared embeddings [4]. Additionally, techniques like imposing commutativity [5] or symmetry [6] on state operators can be utilized to enhance the modeling process.

CZSL requires a specific dataset for training, which should include objects in a variety of transformed states, such as a white dog, a big dog, or an old dog. This diverse dataset enables CZSL to learn how states transform objects. However, during the training of CZSL, the search space is limited to a subset of compositions present in the dataset. If the search space is expanded to include all possible compositions, it is referred to as an open world environment. It should be noted that expanding the search space to the open world environment may lead to a decrease in the accuracy of CZSL. Taking the MIT-states dataset as an example, which consists of 115 attributes and 245 objects, there are a total of 28,175 potential combinations. In the CZSL method, only 1662 components of the search space are observed in the dataset, accounting for less than 6% of all possible components. This restriction of the search space to a subset of compositions exemplifies the close world environment in CZSL.

The third method is *Open-World Compositional Zero Shot Learning* (OW-CZSL), which builds upon the foundations of SCL by introducing an additional interaction model. In CZSL, expanding the search space to include all possible compositions is referred to as an open world environment. This expansion is driven by the need to adapt CZSL to handle a wider range of compositions and accommodate unseen compositions. However, expanding the search space in CZSL leads to a significantly larger output space, which poses challenges in terms of computational complexity and scalability. To address this issue, OW-CZSL takes a different approach by separately identifying states and objects, effectively reducing the output space. This separation allows for more efficient modeling and inference processes. Unlike SCL, OW-CZSL introduces an additional model specifically designed to capture the interaction between states and objects. This interaction model is trained to establish the feasibility between different states and objects. For instance, some studies utilize external knowledge to train a feasibility model that determines the acceptability of a composition [3]. The recognition results from the state model and object model are combined with the feasibility scores to make the final judgment.

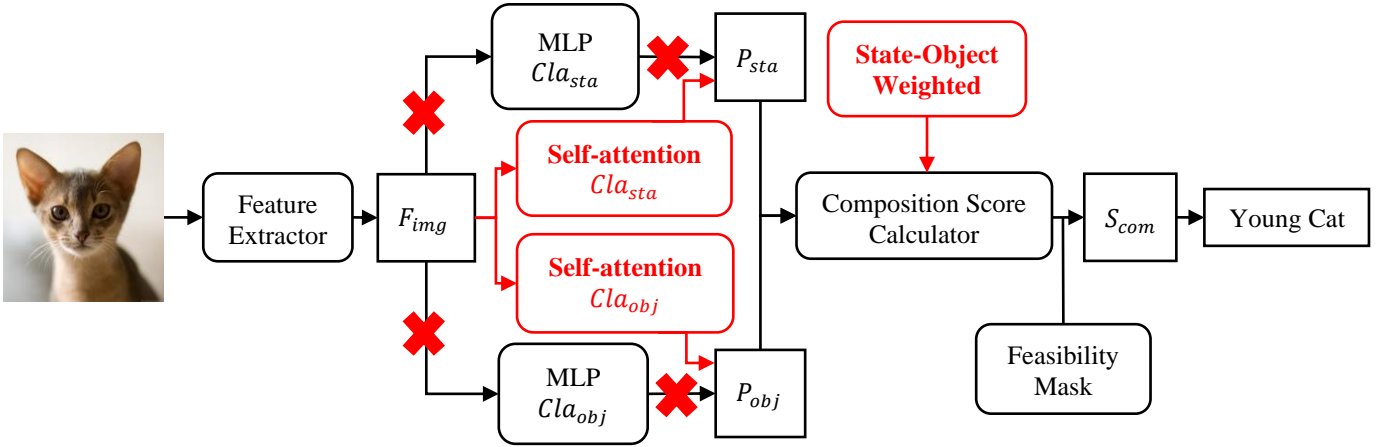

**Fig. 2.** The recognition process of OW-CZSL is depicted in the diagram, where the red-colored components highlight the improvements made in this study. Specifically, in the classifier section, the concept of self-attention is incorporated. Additionally, the Composition score calculator incorporates the notion of weighted combination.

Among the three approaches, OW-CZSL is the mainstream focus of research. Compared to OW-CZSL, SCL does not consider the interaction between states and objects, leading to unreasonable compositions while CZSL, on the other hand, only focuses on compositions present in the dataset, without considering the diversity and generality of compositions. OW-CZSL combines the strengths of both SCL and CZSL while addressing their shortcomings. OWCZSL employs multiple models to reduce the computational burden of directly recognizing compositions, while also accounting for the interaction between state-objects and the broad range of compositions.

Furthermore, the architecture of OW-CZSL is well-suited for recognizing multi-state objects. Based on OW-CZSL, a framework has been designed for identifying multi-state objects. States can be categorized into different types, such as color, shape, and material. Each type employs a distinct classifier for recognition. Subsequently, the framework considers whether the states can form reasonable compositions with objects and even takes into account potential conflicts among different states. However, due to the current limitations of datasets, only single-state object datasets are predominantly used for experimentation and discussion in this study.

The current state-of-the-art OW-CZSL based approach is KG-SP. KG-SP first employs ResNet to extract image features and utilizes two MLP classifiers to predict states and objects respectively. The results from these classifiers are used to calculate composition scores. The feasibility mask generated by a semantic model is then utilized to filter rational compositions. Nevertheless, we have observed that the classifiers used in KG-SP yield inferior results when dealing with complex images, such as those with varying object positions or intricate backgrounds. Additionally, during the computation of composition scores, KG-SP does not account for the differences between states and objects, treating them with equal weights when combining them. However, in reality, the classifiers for states and objects do not consistently generate the same level of confidence, implying varying degrees of influence when forming compositions.

To solve the problems, we propose *Self-Attention with State-Object Weighted Combination* (SASOW) to improve KG-SP, as depicted in Figure 2. This enhancement involves incorporating classifiers into a self-attention mechanism and modifying the original MLP architecture. This adaptation enhances the recognition capability of classifiers for complex images. Moreover, during the computation of combination scores, the framework takes into account the disparities between states and objects. Different weights are assigned based on the confidence levels of the two classifiers. This differential weighting allows states and objects to exert distinct influences on compositions, thereby generating more accurate combination outcomes.

The remainder of this paper is structured as follows: Section II provides an introduction to the related works in the field of CZSL and OW-CZSL, as well as an overview of KG-SP, which serves as the foundation for our proposed SASOW. In Section III, we present an overview of SASOW, detailing the Self-Attention mechanism applied to the classifier and the weight combination approach. The experimental results and ablation studies are presented in Section IV. Finally, in Section V, we provide a summary of the conclusions drawn from this work and discuss potential avenues for future work.

II. RELATED WORK

*A. Compositional Zero Shot Learning*

The objective of *Compositional Zero Shot Learning* (CZSL) is to recognize state-object compositions in images, even for compositions that have not been encountered during training. Modeling the transformation of states to objects is a crucial task in CZSL, and one viable approach for modeling this is through classifier combination. This entails merging individual concept classifiers into a composite concept classifier using techniques such as tensor completion, Boolean algebra, or linear transformation.

Alternatively, state transformations can be represented by simulating them as transformations in vector spaces. Common CZSL methods include *Attributes as Operators* (AoP) introduced by Nagarajan et al. [5], where representations of unseen state-object compositions are generated by applying the corresponding state transformation matrix to the vector of target objects. *Task Modular Networks* (TMN) [7] consist of a set of small fully connected layers operating in the semantic concept space. These modules are configured using a task-dependent gating function to generate features representing the compatibility between input images and the considered concepts. *Compositional Graph Embeddings* (CGE) [8] leverage the dependencies among states, objects, and compositions in a graph structure [24] to facilitate the transfer of relevant knowledge between them. *SymNet* [6] exploits the principle of symmetry to enhance the modeling of the transformation process.

However, during CZSL training, the search space is typically limited to a subset of compositions present in the dataset. Expanding the search space to encompass all possible compositions in an open-world setting can lead to a decline in accuracy. To address this challenge and extend the search space to an open-world scenario, the OW-CZSL method is proposed by *Compositional Cosine Logits* (CompCos) [9].

*B. Knowledge Guided Simple Primitives (KG-SP)*

In *Open-World Compositional Zero-Shot Learning* (OW-CZSL), there are no priors available for unseen compositions, and models need to consider all possible compositions during testing. Due to the large cardinality of the output space, it is challenging to generate discriminative embeddings for the unseen compositions [9]. Inspired by the findings of [9], this work explores a completely different approach. Specifically, an architecture is designed that disregards the compositional nature of the problem and independently produces initial predictions for objects and states. The rationale is that while discriminating between compositions is difficult in OW-CZSL due to the extensive search space, recognizing primitives (i.e., objects and states) in isolation is easier because 1) the cardinality of the two sets is much smaller, and 2) the sets remain fixed during both training and testing.

Drawing inspiration from [1] and [2], a simple method is devised to predict objects and states using two independent classifiers. Recognizing states requires different features compared to recognizing objects, so instead of sharing a feature representation, the model is trained with two distinct nonlinear feature extractors. Moreover, since not all compositions are equally feasible in reality (e.g., a ripe dog), the model's predictions can be refined by removing less feasible compositions from the output space. To achieve this, external knowledge from *ConceptNet* [16] is utilized to estimate the compatibility between a state and an object, allowing the removal of less feasible compositions during testing. The proposed model is named *Knowledge-Guided Simple Primitives* (KG-SP).

*C. Self-attention*

Self-attention is a technique employed in handling sequential data [12]. The fundamental principle of this technique is to regard each element within a sequence as comprising a query, a key, and a value. Similarities among these components are computed to derive weighted sums, resulting in representative vectors for these elements. Specifically, for each query, the similarity to all keys is computed, and the resultant weights are utilized to perform weighted summation over all corresponding values. In this manner, a representative vector is obtained for each element in the sequence, emphasizing interactions with all other elements. Self-attention has found widespread application across various domains, including the realm of images [13]. Notably, the *Vision Transformer* (ViT) model utilizes self-attention as its cornerstone, representing an image classification model rooted in the self-attention mechanism. The ViT model dissects images into smaller patches, treating them as a sequence for input into the Transformer model. In the course of processing, the self-attention mechanism assists the model in discerning relationships between distinct regions, thereby enhancing the model's classification performance.

The efficacy of self-attention in the realm of images lies in its ability to capture the interrelationships between different parts of an image. In tasks involving image classification and recognition, the model needs to comprehend the interactions among various components of the image to better differentiate categories. This is particularly crucial for compositional learning, as discerning states and objects requires a heightened focus on the relationships between individual features. This adaptability becomes especially significant when dealing with intricate images, such as those with varying object positions or complex backgrounds. By incorporating self-attention, the model can allocate greater attention to the relationships between different regions within the image. Consequently, it becomes more adept at capturing image features effectively and enhancing its overall performance. This, in turn, elevates its capability to recognize complex images. While this aspect has not been thoroughly discussed in comparison to other methods in the CZSL, it stands as a pivotal factor in augmenting the recognition abilities of states and objects.

III. METHODOLOGY

This study aims to achieve effective performance in an open world environment by building upon the state-of-the-art method KG-SP [3] in OW-CZSL. We employ two classifiers, one for recognizing states and the other for identifying objects. To ensure the feasibility of the generated combinations, we utilize a feasibility mask obtained through training from Conceptnet. The optimized framework, as depicted in Fig 1, involves passing the image through a feature extractor, followed by the state classifier and object classifier. In order to focus on specific features during state and object recognition, we incorporate Self-Attention mechanisms into both classifiers. After the recognition phase, when composing the scores for states and objects, we introduce State-Object weighting to bias the combination towards the more accurate component, leading to more precise results.

*A. Self-Attention on Classifier*

The proposed framework in KG-SP initially utilizes Resnet for extracting image features, capturing both local and global characteristics of the images. The resulting image feature vector is referred to as $F_{img}$. Subsequently, this feature vector passes through the state and object classifiers, $Cla_{sta}$ and $Cla_{obj}$, respectively. These classifiers are constructed using a simple *Multi-Layer Perceptron* (MLP) [21] architecture, as shown in Fig 2. To enhance the precision of both state and object recognition, we have introduced modifications to the original MLP architecture by incorporating a Self-Attention mechanism, as depicted in Fig. 3. After passing through a Linear layer, $F_{img}$ is separately fed into three Linear layers representing *Query* (Q), *Key* (K), and *Value* (V). The vectors resulting from the multiplication of K and Q are transformed into probability distributions, known as attention scores. These attention scores are then multiplied with the vectors from the V layer, generating a set of scores called attention weights. Finally, the attention weights are added to the original $F_{img}$, yielding a set of attention features that have been enhanced through the Self-Attention mechanism.

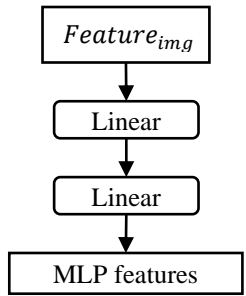

**Fig. 2.** The structure of original MLP on classifier.

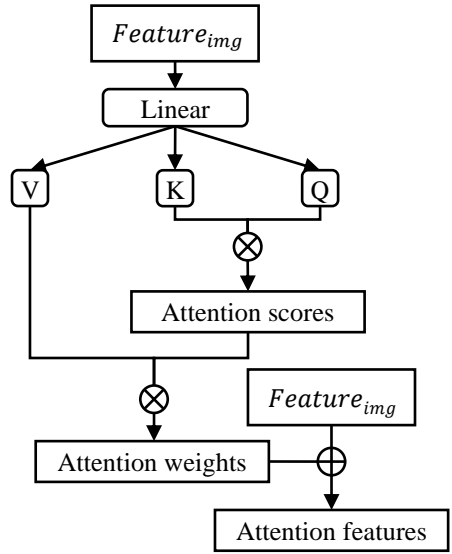

**Fig. 3.** The structure of self-attention mechanism on classifier.

TABLE I UNIT FOR COMPOSITIONAL LEARNING

| Symbol | Description |
|---|---|
| $F_{img}$ | Features obtained through ResNet for images. |
| $Cla_{sta}$ | Classifier used for classifying state. |
| $Cla_{obj}$ | Classifier used for classifying object. |
| $P_{sta}$ | The probability scores generated by $Cla_{sta}$ represent the probability of each state being the state in the image. |
| $P_{obj}$ | The probability scores generated by $Cla_{obj}$ represent the probability of each object being the object in the image. |
| $A_{sta}$ | Accuracy based only on the state portion. |
| $A_{obj}$ | Accuracy based only on the object portion. |
| $S_{com}$ | Composition score obtained through the combination of $P_{sta}$ and $P_{obj}$ |

By incorporating the Self-Attention mechanism at the feature level, this modified model effectively models the interactions and importance within the input sequence, enabling better capture of the relationships between features. This attention mechanism allows the model to focus on important features and adapt to input relationships across different positions. Compared to the original simple MLP model, the modified model demonstrates enhanced expressive and modeling capabilities when dealing with complex sequential data.

## B. Weighted Combination

When the $F_{img}$ passes through the $Cla_{sta}$ and $Cla_{obj}$, two sets of vectors that correspond to unique one-dimensional matrices will be obtained, one of the vectors $P_{sta}$, which represents the probability distribution that the image may belong to each state after passing through the $Cla_{sta}$, and the other vector, $P_{obj}$, represents the probability distribution that the picture may belong to each object after passing through the $Cla_{obj}$. The lengths of these two vectors are the classification state and the number of types of objects respectively. For example, if our dataset contains 2 states and 3 object types, we will obtain a state vector with 2 elements, $[P_{sta_1}, P_{sta_2}]$, and an object vector with 3 elements, $[P_{obj_1}, P_{obj_2}, P_{obj_3}]$.

To merge the resulting vectors $P_{sta}$ and $P_{obj}$ generated by $Cla_{sta}$ and $Cla_{obj}$, respectively, we first transpose the object vector to create a state row vector and an object column vector. Then, we utilize the method of *batch matrix multiplication* [25][26] to combine the two sets of vectors, as shown in the following equation:

$$S_{com_i} = P_{sta_i} @ P_{obj_i}$$

In this method, the corresponding elements of the two vectors are multiplied to obtain the score for each composition of state and object, indicating the likelihood of the image belonging to that particular composition. The reason for using batch matrix multiplication is to multiply two independent probabilities in a manner similar to joint probability distribution, resulting in the probability distribution of each composition of state and object, denoted as $S_{com}$, represented as a one-dimensional matrix. Continuing with the previous example, we can perform batch matrix multiplication between the state vector and the object vector to obtain six combination scores $[P_{sta_1}P_{obj_1}, P_{sta_1}P_{obj_2}, P_{sta_1}P_{obj_3}, P_{sta_2}P_{obj_1}, P_{sta_2}P_{obj_2}, P_{sta_2}P_{obj_3}]$, representing the likelihood of the image belonging to each state and object combination.

However, since the object classifier performance is usually better than the state classifier, in order to improve the accuracy of the combination, all elements $P_{sta_i}$ in $P_{sta}$ are exponentiated with the ratio of the accuracy of $Cla_{sta}$, $A_{sta}$, to the accuracy of $Cla_{obj}$, $A_{obj}$. This results in a new matrix vector $P_{sta}'$, as shown in the following equation:

$$P_{sta_i}' = (P_{sta_i})^{\frac{A_{sta}}{A_{obj}}}$$

Finally, the batch matrix multiplication is performed between $P_{sta}'$ and $P_{obj}$, resulting in the following equation:

$$S_{com_i}' = P_{sta_i}' @ P_{obj_i}$$

Since the recognition effect of the object classifier is usually better, the ratio $\frac{A_{sta}}{A_{obj}}$ is usually smaller than 0, which will reduce the gap between the lower probability and the higher probability in $P_{sta}$ while maintaining the size relationship. Hence, $P_{obj}$ will contribute more to the $S_{com}$, while $P_{sta}$ will contribute less to the $S_{com}$. This can increase our confidence in the object classification effect and further improve the accuracy of the composition.

## IV. EXPERIMENTS

### A. Dataset

Three datasets are employed for conducting open-world compositional zero-shot learning: MIT-States [17], UT-Zappos [18,19], and C-GQA [7]. MIT-States is a large-scale dataset encompassing vastly diverse states and objects, ranging from items to animals. It comprises 115 state categories and 245 object categories, resulting in a total of 1662 actual compositions. UT-Zappos, on the other hand, consists of shoe images, comprising 12 states, 16 objects, resulting in 101 compositions. C-GQA is derived from GQA and incorporates 413 states, 674 objects, resulting in 7985 compositions.

TABLE II DATASET DETAILS

| Dataset | Train set | | | Test set | |
|---|---|---|---|---|---|
| | SC | i | SC | UC | i |
| MIT-States | 1262 | 22998 | 400 | 400 | 2914 |
| UT Zappos | 83 | 30338 | 18 | 18 | 12995 |
| C-GQA | 6963 | 26920 | 1022 | 1047 | 5098 |

The datasets are divided into training and testing sets. Certain compositions within the datasets are designated as *seen compositions* (SC), utilized for training purposes. Conversely, compositions not falling under the category of seen compositions are termed *unseen compositions* (UC) and are reserved solely for testing. The testing sets comprise both seen and unseen compositions, enabling separate evaluation of the model's performance on each. Detailed partitioning data for the three datasets is presented in Table II, where "i" corresponds to the respective number of images. All subsequent experiments are based on this train-test split.

*B. Baseline*

Our approach is compared against standard CZSL methods in an open-world environment. These methods include CZSL, *Attribute as Operators* (AoP) [5], *Label Embedding+* (LE+) [1], *Task Modular Network* (TMN) [4], *SymNet* [6], *Compositional Graph Embedding* (CGE) [7], and OW-CZSL methods, namely *Compositional Cosine Logits* (CompCos) [9] and the state-of-the-art method, *Knowledge-Guided Simple Primitives* (KG-SP) [3]. These methods are widely discussed and referenced within the CZSL domain.

*C. Evaluation metrics*

We adopt the standard splits of [4,7] and evaluate all the methods under the generalized setting, where the model is tested on samples from both seen and unseen compositions. Following the protocol outlined in [4], we assess the performance based on two key metrics: the best *seen accuracy* (S) and the best *unseen accuracy* (U). The best seen accuracy represents the accuracy on the test set samples belonging to seen compositions, while the best unseen accuracy represents the accuracy on the test set samples belonging to unseen compositions. Furthermore, we calculate the best harmonic mean (HM), which is the harmonic mean of S and U, as well as the area under the curve (AUC) for both S and U. These metrics provide a comprehensive understanding of the method's performance in composition recognition, with improvements in U indicating the method's ability to capture the interaction between states and objects.

*D. Result*

The performance of SASOW on the challenging benchmark OW-CZSL is discussed in this paragraph. Table III presents the results of our experiments, where SASOW is compared to state-of-the-art methods in the field. According to the findings, SASOW either surpasses or competes with the state-of-the-art methods in all metrics on the OW-CZSL setting. Specifically, on the C-GQA dataset, SASOW achieved the best results across all metrics. In comparison to the state-of-the-art method KG-SP, SASOW demonstrated improvements of 1.1% (32.7 vs 31.5) in best seen accuracy, 0.4% (3.3 vs 2.9) in best unseen accuracy, 0.3% (5.0 vs 4.7) in best harmonic mean, and 0.07% (0.85 vs 0.78) in best AUC. Additionally, SASOW exhibited competitive performance on the MIT-States and UT Zappos datasets.

Furthermore, we conducted comparisons of SASOW and other methods under the same configuration and environment, resulting in improvements across all three datasets with SASOW. This validates the effectiveness of the techniques employed by SASOW in enhancing CZSL. These findings reinforce the significance of SASOW as a valuable approach in the CZSL domain. Moreover, this experiment confirms the superiority of SASOW over other advanced methods like KG-SP. These results provide crucial support for further exploration and development of novel approaches in the CZSL field.

TABLE III OPEN WORLD CZSL RESULTS ON MIT-STATES, UT ZAPPOS AND C-GQA. WE MEASURE BEST SEEN (S) AND UNSEEN ACCURACY (U), BEST HARMONIC MEAN (HM), AND AREA UNDER THE CURVE (AUC) ON THE COMPOSITIONS.

| Method | MIT-States | | | | UT Zappos | | | | C-GQA | | | |
|---|---|---|---|---|---|---|---|---|---|---|---|---|
| | S | U | HM | AUC | S | U | HM | AUC | S | U | HM | AUC |
| TMN | 12.6 | 0.9 | 1.2 | 0.1 | 55.9 | 18.1 | 21.7 | 8.4 | NA | NA | NA | NA |
| AoP | 16.6 | 5.7 | 4.7 | 0.7 | 50.9 | 34.2 | 29.4 | 13.7 | NA | NA | NA | NA |
| LE+ | 14.2 | 2.5 | 2.7 | 0.3 | 60.4 | 36.5 | 30.5 | 16.3 | 19.2 | 0.9 | 1.0 | 0.08 |
| VisProd | 20.9 | 5.8 | 5.6 | 0.7 | 54.6 | 42.8 | 36.9 | 19.7 | 24.8 | 1.7 | 2.8 | 0.33 |
| SymNet | 21.4 | 7.0 | 5.8 | 0.8 | 53.3 | 44.6 | 34.5 | 18.5 | 26.7 | 2.2 | 3.3 | 0.43 |
| CompCos | 25.4 | **10.0** | **8.9** | **1.6** | 59.3 | 46.8 | 36.9 | 21.3 | 28.4 | 1.8 | 2.8 | 0.39 |
| CGE | **32.4** | 5.1 | 6.0 | 1.0 | 61.7 | 47.7 | 39.0 | 23.1 | **32.7** | 1.8 | 2.9 | 0.47 |
| KG-SP(paper) | 28.4 | 7.5 | 7.4 | 1.3 | 61.8 | 52.1 | **42.3** | **26.5** | 31.5 | 2.9 | 4.7 | 0.78 |
| KG-SP (Implementation) | 26.7 | 7.8 | 7.6 | 1.3 | 62.6 | 51.5 | 40.9 | 25.7 | 31.6 | 2.7 | 4.3 | 0.69 |
| SASOW (Ours) | 27.4 | 9.6 | 8.6 | **1.6** | **63.5** | **53.1** | 40.2 | 25.2 | **32.7** | **3.3** | **5.0** | **0.85** |

In addition to evaluating composition recognition, we also separately evaluate the accuracy of states (Sta.) and objects (Obj.). This allows us to gain insights into the performance of the method in recognizing states and objects, which is crucial for understanding the comparative effectiveness of different approaches and providing valuable references for weight adjustments. In Table IV, we compare SASOW with the current state-of-the-art method KG-SP. Due to the incorporation of weight combination in SASOW, its accuracy in recognizing states is lower than KG-SP, while its accuracy in recognizing objects is higher than KG-SP. However, this also indicates that under this particular combination of states (Sta.) and objects (Obj.), more accurate compositions can be formed.

TABLE IV STATE AND OBJECT RESULTS ON MIT-STATES, UT ZAPPOS AND C-GQA.

| Method | MIT-States | | UT Zappos | | C-GQA | |
|---|---|---|---|---|---|---|
| | Sta. | Obj. | Sta. | Obj. | Sta. | Obj. |
| KG-SP | 19.9 | 29.9 | 51.5 | 72.6 | 29.8 | 38.2 |
| SASOW | 19.8 | 32.6 | 49.1 | 74.4 | 22.4 | 43.2 |

*D. Ablation Study*

In this subsection, we discuss the results of ablation experiments conducted on SASOW. We incorporated self-attention into KG-SP, denoted as KG-SA, and introduced weighted combination into KG-SP, termed KG-SOW. Table IV presents a comparison of these two approaches with KG-SP and SASOW. From Table IV, it is evident that self-attention yields more pronounced effects on MIT-States and C-GQA datasets, while weighted combination demonstrates more favorable outcomes on the UT Zappos dataset.

*1) Self-attention*

From Table V, it can be observed that self-attention is more effective for the MIT-States and C-GQA datasets compared to UT Zappos. This is likely due to the complexity of the MIT-States and C-GQA datasets, where objects are distributed in various positions within the image and the background is more intricate. In contrast, UT Zappos contains objects positioned centrally in the image and has a uniform white background. We hypothesize that the classifier in KG-SP may have difficulty in locating objects accurately.

To test this idea, we conducted an experiment where we randomly moved objects within UT Zappos images and replaced the background with random colors (as shown in Fig. 4 during training. We call this new dataset UT Zappos-moving. This was done to create a more challenging scenario where object positions are not fixed and the background becomes more complex and varied. By introducing this random displacement and background alteration, we aimed to evaluate the impact of self-attention in improving the model's ability to accurately locate objects in such challenging conditions.

TABLE VI THE RESULTS AFTER TRAINING WITH UT ZAPPOS-MOVING.

| Method | UT Zappos-moving | | | | | |
|---|---|---|---|---|---|---|
| | S | U | HM | AUC | Sta. | Obj. |
| KG-SP | 55.6 | 45.2 | 35.9 | 20.1 | 44.5 | 71.4 |
| KG-SA | 61.3 | 47.1 | 36.9 | 21.8 | 45.6 | 71.9 |

TABLE V THE RESULTS OBTAINED AFTER ADDING SELF-ATTENTION AND WEIGHT COMBINATION TO TRAINING AFTER EACH DATA SET.

| Method | MIT-States | | | | UT Zappos | | | | C-GQA | | | |
|---|---|---|---|---|---|---|---|---|---|---|---|---|
| | S | U | HM | AUC | S | U | HM | AUC | S | U | HM | AUC |
| KG-SP | 26.7 | 7.8 | 7.6 | 1.3 | 62.6 | 51.5 | 40.9 | 25.7 | 31.6 | 2.7 | 4.3 | 0.69 |
| KG-SA | 27.2 | 9.8 | 8.8 | 1.6 | 63.5 | 50.8 | 40.2 | 25.4 | 32.8 | 3.3 | 4.9 | 0.84 |
| KG-SOW | 26.8 | 7.7 | 7.5 | 1.3 | 62.6 | 52.1 | 40.8 | 25.5 | 31.4 | 3.0 | 4.7 | 0.72 |
| SASOW (Ours) | 27.4 | 9.6 | 8.6 | 1.6 | 63.5 | 53.1 | 40.2 | 25.2 | 32.7 | 3.3 | 5.0 | 0.85 |

We compared the performance with and without self-attention, and the results in Table VI demonstrate improvements in all evaluation metrics after incorporating self-attention. This confirms that self-attention can enhance the performance on datasets where object positions are not fixed and backgrounds are complex.

*2) Weighted combination*

We conducted additional experiments on the UT Zappos dataset to compare the training results with and without weight combination under different numbers of training compositions. The number of images in the test set was fixed, and the test set included the same set of seen and unseen compositions. The variable part was the compositions that existed only in the train set, which we referred to as "only train compositions". We gradually reduced the number of these "only train compositions" in the train set while maintaining the setting of selecting one answer from all possible compositions in an open-world environment. This ensured that the training set still encompassed all the states and objects. We named the datasets with different numbers of training compositions and recorded their details in Table VII, where "*com*" represents the number of compositions included and "*i*" represents the number of images included. Based on the results in Table VIII, it can be observed that weight combination maintained the accuracy for seen compositions and effectively improved the accuracy for unseen compositions, regardless of the number of training compositions used.

TABLE VII

DATASET STATISTICS FOR THE TRAINING SET WITH VARYING NUMBERS OF TRAIN COMPOSITIONS.

| Dataset | Train | | | |
|---|---|---|---|---|
| | All | | Seen | |
| | *com* | *i* | *com* | *i* |
| UT Zappos | 83 | 23875 | 18 | 4131 |
| UT60 | 60 | 16234 | 18 | 4131 |
| UT45 | 45 | 10377 | 18 | 4131 |
| UT36 | 36 | 8414 | 18 | 4131 |

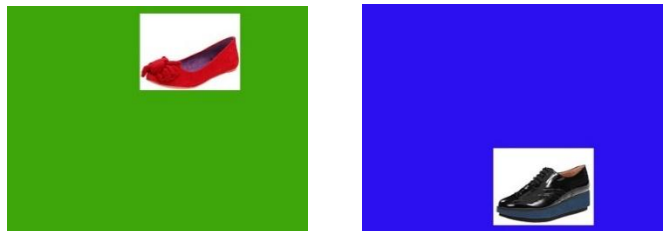

Fig. 4. In order to verify the effect of self-attention, we randomly move the pictures in UT Zappos and change the background to random colors.

## V. CONCLUSION

In this paper, a novel OW-CZSL framework named SASOW has been introduced, which has been adapted to multiple states. Firstly, self-attention was integrated into both the state classifier and object classifier, enabling the model to proficiently handle complex images, thereby enhancing recognition accuracy in composite scenarios. Furthermore, when merging the results of the state classifier and object classifier, consideration has been given to the varying confidences of these classifiers, and a weighted combination approach was employed to formulate more coherent composite scores. It is worth noting that SASOW has demonstrated competitive performance on three single-state object datasets (MIT-States, UT Zappos, and C-GQA) with best unseen accuracy reaching 9.6%, 53.1%, and 3.3%, respectively. Compared to KG-SP, this represents an improvement of 2.1%, 1.7%, and 0.4%. The significant reason behind these results is that SASOW is more adaptable to complex images and computes composition scores that closely approximate real-world scenarios, all achieved in the present perfect tense.

With the continuous advancement of computer vision, the potential applications of CZSL have significantly increased in various domains. In order to truly extend the applicability of CZSL methods to different fields, it is essential to establish domain-specific datasets and evaluate the feasibility of applying these methods, thereby adapting them to different domain-specific datasets. Furthermore, it is worth noting that objects can exist in multiple states, which makes the study of multi-state scenarios a future challenge. For example, defining labeling schemes for multi-states and developing methods to evaluate the recognition of multi-state combinations can be explored. Additionally, when recognizing multi-state combinations, it can be determined whether the states of different categories are mutually contradictory, and the influence of different categories on the final combination can be considered using the weighted combination approach proposed in this paper. Apart from these directions, exploring different methods for extracting image features is crucial. This can further improve the accuracy of state and object recognition and allow for the more effective utilization of self-attention mechanisms. In summary, developing methods to recognize multi-state compositions and improving the accuracy of object and state recognition are two directions that can be further explored in the future.

TABLE VIII

RESULTS OF WHETHER WEIGHT COMBINATION WAS EMPLOYED DURING TRAINING WITH DATASETS CONTAINING DIFFERENT NUMBERS OF TRAIN COMPOSITIONS.

| Dataset | Method | Unseen | Seen | State | Object |
|---|---|---|---|---|---|
| UT Zappos | KG-SA | 50.8 | 63.5 | 51.2 | 72.1 |
| | SASOW | **53.1** | 63.5 | 49.1 | 74.4 |
| UT60 | KG-SA | 34.8 | 68.4 | 42.1 | 67.5 |
| | SASOW | **35.4** | 68.5 | 39.7 | 71.2 |
| UT45 | KG-SA | 20.1 | 74.4 | 35.3 | 62.9 |
| | SASOW | **21.2** | 74.5 | 34.8 | 68.9 |
| UT36 | KG-SA | 11.8 | 80.0 | 33.1 | 61.0 |
| | SASOW | **14.0** | 79.8 | 32.3 | 67.9 |

REFERENCES

[1] Ishan Misra, Abhinav Gupta, and Martial Hebert, " From red wine to red tomato: Composition with context," In CVPR, 2017.

[2] Christoph H Lampert, Hannes Nickisch, and Stefan Harmeling," Learning to detect unseen object classes by between-class attribute transfer," In CVPR, 2009.

[3] Karthik, S and Mancini, M and Akata, Zeynep, " KG-SP: Knowledge Guided Simple Primitives for Open World Compositional Zero-Shot Learning," 35th IEEE Conference on Computer Vision and Pattern Recognition,2022.

[4] Senthil Purushwalkam, Maximilian Nickel, Abhinav Gupta, and Marc'Aurelio Ranzato, " Task-driven modular networks for zero-shot compositional learning," In ICCV, 2019.

[5] Tushar Nagarajan and Kristen Grauman, " Attributes as operators: factorizing unseen attribute-object compositions," In ECCV, 2018.

[6] Yong-Lu Li, Yue Xu, Xiaohan Mao, and Cewu Lu, " Symmetry and group in attribute-object compositions," In CVPR, 2020.

[7] Muhammad Ferjad Naeem, Yongqin Xian, Federico Tombari, and Zeynep Akata," Learning graph embeddings for compositional zero-shot learning," In CVPR, 2021.

[8] Frank Ruis, Gertjan J Burghouts, and Doina Bucur, " Independent prototype propagation for zero-shot compositionality," In NeurIPS, 2021.

[9] Massimiliano Mancini, Muhammad Ferjad Naeem, Yongqin Xian, and Zeynep Akata, "Open world compositional zero shot learning," In CVPR, 2021.

[10] C.-Y. Chen and K. Grauman, "Inferring analogous attributes," in CVPR, 2014.

[11] R. S. Cruz, B. Fernando, A. Cherian, and S. Gould, "Neural algebra of classifiers," In WACV, 2018.

[12] Ashish Vaswani, Noam Shazeer, Niki Parmar, Jakob Uszkoreit, Llion Jones, Aidan N. Gomez, Lukasz Kaiser, Illia Polosukhin,"Attention Is All You Need" In NIPS, 2017.

[13] Alexey Dosovitskiy, Lucas Beyer, Alexander Kolesnikov, Dirk Weissenborn, Xiaohua Zhai, Thomas Unterthiner, Mostafa Dehghani, Matthias Minderer, Georg Heigold, Sylvain Gelly, Jakob Uszkoreit, Neil Houlsby, "An Image is Worth 16x16 Words: Transformers for Image Recognition at Scale," In ICLR,2021

[14] D. Comite and N. Pierdicca, "Decorrelation of the near-specular land scattering in bistatic radar systems," IEEE Trans. Geosci. Remote Sens., early access, doi: 10.1109/TGRS.2021.3072864. (Note: This format is used for articles in early access. The doi must be included.)

[15] H. V. Habi and H. Messer, "Recurrent neural network for rain estimation using commercial microwave links," IEEE Trans. Geosci. Remote Sens., vol. 59, no. 5, pp. 3672-3681, May 2021. [Online]. Available: https://ieeexplore.ieee.org/document/9153027

[16] Robyn Speer, Joshua Chin, and Catherine Havasi. "Conceptnet 5.5: An open multilingual graph of general knowledge." In AAAI, 2017.

[17] Phillip Isola, Joseph J Lim, and Edward H Adelson. Discovering states and transformations in image collections. In CVPR, 2015.

[18] Aron Yu and Kristen Grauman. Fine-grained visual comparisons with local learning. In CVPR, 2014.

[19] Aron Yu and Kristen Grauman. Semantic jitter: Dense supervision for visual comparisons via synthetic images. In ICCV, 2017.

[20] Kaiming He, Xiangyu Zhang, Shaoqing Ren, and Jian Sun. Deep residual learning for image recognition. In CVPR, 2016.

[21] F. Murtagh, "Multilayer perceptrons for classification and regression," Neurocomputing, vol. 2, nos. 5–6, pp. 183–197, 1991.

[22] Devi Parikh and Kristen Grauman. "Relative attributes," In ICCV, 2011.
[23] Ali Farhadi, Ian Endres, Derek Hoiem, and David Forsyth. "Describing objects by their attributes," In CVPR, 2009.
[24] Thomas N. Kipf, Max Welling, "Semi-supervised classification with graph convolutional networks," In ICLR, 2017.
[25] C. Lu, R. Krishna, M. Bernstein, and L. Fei-Fei. "Visual relationship detection with language priors," In ECCV, 2016
[26] Paszke, A., Gross, S., Massa, F., Lerer, A., Bradbury, J., Chanan, G., … Chintala, S. (2019). "PyTorch: An Imperative Style, High-Performance Deep Learning Library," In Advances in Neural Information Processing Systems 32 (pp. 8024–8035).